\documentclass{article}

\usepackage[final]{nips_2017}

\usepackage[utf8]{inputenc} 
\usepackage[T1]{fontenc}    
\usepackage{hyperref}       
\usepackage{url}            
\usepackage{booktabs}       
\usepackage{amsfonts}       
\usepackage{nicefrac}       
\usepackage{microtype}      

\usepackage{natbib}

\usepackage{graphicx,color}
\usepackage{amsmath,amsthm,amssymb}
\usepackage{subfigure}
\usepackage{multirow}
\usepackage{enumitem}
\usepackage[linesnumbered,ruled]{algorithm2e}

\SetCommentSty{mycommfont}

\theoremstyle{remark}

\usepackage[flushleft]{threeparttable}

\usepackage{tabulary}
\graphicspath{../Figures}

\title{A Forward-Backward Approach for Visualizing Information Flow in Deep Networks}

\author{
  Aditya Balu, Thanh V. Nguyen, Apurva Kokate, Chinmay Hegde, Soumik Sarkar\thanks{\{baditya,thanhng,akokate,chinmay,soumik*@iastate.edu\}.} \\
   Iowa State University\\
  Ames, IA 50010 \\
}

\begin{document}

\maketitle

\begin{abstract}

We introduce a new, systematic framework for visualizing information flow in deep networks. Specifically, given any trained deep convolutional network model and a given test image, our method produces a compact support in the image domain that corresponds to a (high-resolution) feature that contributes to the given explanation. Our method is both computationally efficient as well as numerically robust. We present several preliminary numerical results that support the benefits of our framework over existing methods.
\end{abstract}

\section{Introduction}


Deep neural networks have resulted in widespread and compelling advances in a variety of machine learning tasks such as object recognition, image segmentation, anomaly detection, machine translation, and synthesis. However, these advances have often been accompanied by a significant reduction in \emph{interpretability}, or the ability to visualize the flow of information being extracted at various layers of abstraction. In contrast to traditional rule-based learning methods (which search for specific, semantically hand-crafted features or patterns), deep networks often produce decisions that are seemingly hard to decipher or justify for a \emph{given} test data sample, even though their aggregate generalizability measured with respect to a hold-out test dataset is excellent. This issue of unpacking the ``black-box'' nature of deep networks has been identified as a key issue by several recent works~\cite{guided-backprop,gradcam,deeplift,integ-grad}. 

In this work, we focus on the task of object detection in images. Broadly, algorithms for interpreting the action of deep networks for this task can be grouped as follows: Class-discriminative approaches, such as Class Activation Mappings (CAM)~\cite{cam}, or its gradient-based variant~\cite{gradcam}, produce a support in the original image domain that approximately corresponds to a given object class detected in that image. However, such methods are coarse and only produce low-resolution visualizations, and as such cannot be directly applied to very high resolution images. On the other hand, pixel-space gradient-based methods such as deconvolution networks~\cite{deconvnet} and guided back-propagation~\cite{guided-backprop} produce fine-grained features in a given image. However, gradient based methods suffer from either significant computational efficiency concerns, or are susceptible to saturation phenomena due to vanishing/exploding gradients, or both. In~\cite{deeplift}, this issue is alleviated by suitably using a second \emph{reference} input to stabilize the estimates. However, choosing this reference image is qualitative and can be challenging. Finally, model-agnostic approaches such as LIME~\cite{lime} are theoretically sound and can be applied for interpreting deep convolution networks, but involve solving challenging optimization problems.


In this short paper, we outline a systematic framework for visualizing information flow in deep convolutional networks that resolves both the computational efficiency as well as the numerical robustness issues described above. We present several preliminary numerical results that support the benefits of our framework over existing methods. 

\begin{figure}
\centering
\includegraphics[width=0.7\textwidth]{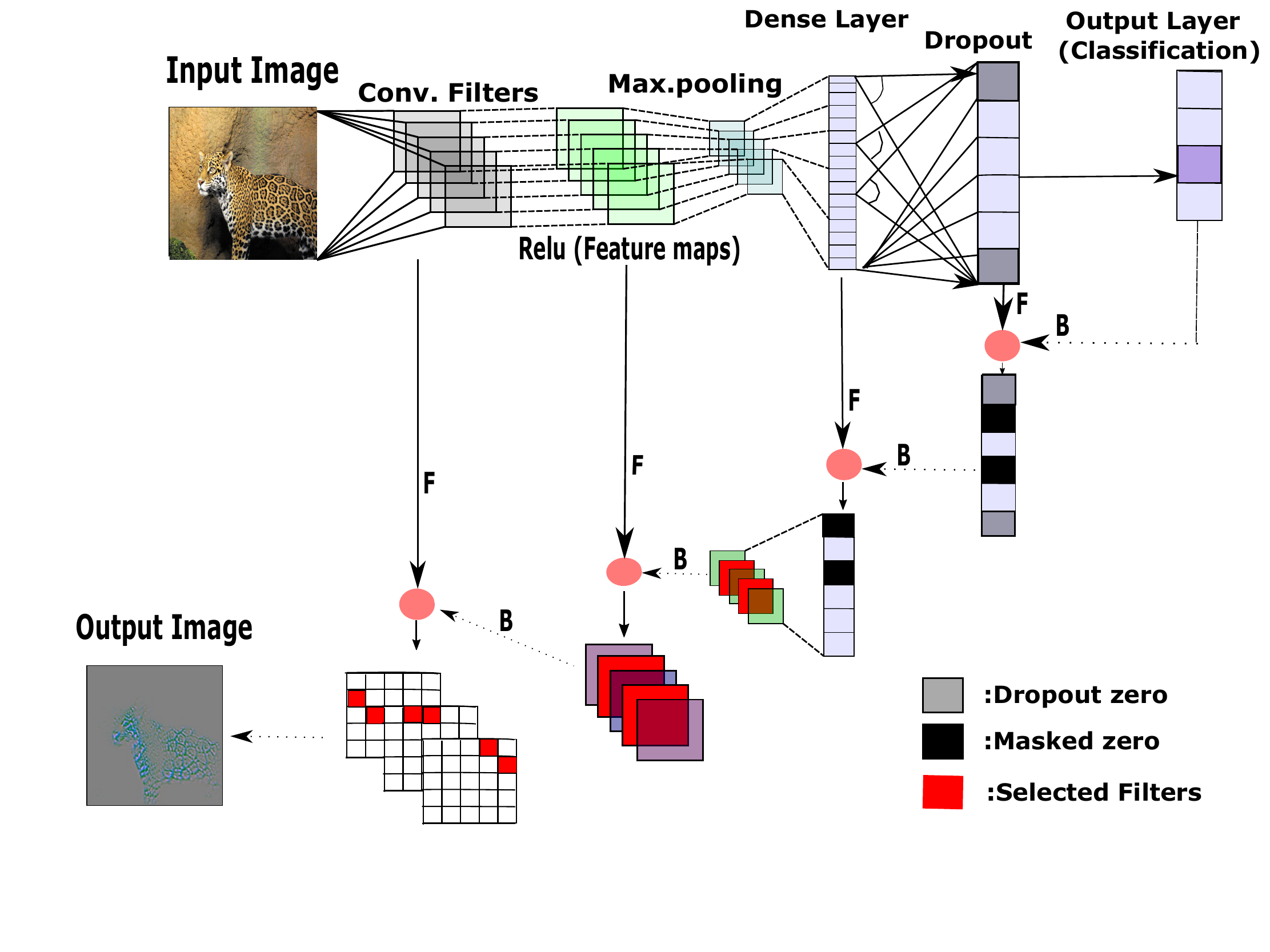}
\caption{\small\sl Overview of the forward-backward scheme for
  visualizing the information flow. Forward information (F) of the
  model is combined with the backward information (B) and flow throughout the network. 
\label{fig:main}}
\end{figure}

At a high level, our approach is based on a novel \emph{forward-backward} scheme which operates as follows. Consider a trained deep convolutional network model and a given test image for which our model is able to identify the existence of a given target class. Then, our algorithm produces as output, a support (i.e., a subset of pixel locations) corresponding to the class predicted by our model in a manner similar to pixel-space gradient methods. However, in contrast with gradient-based approaches, our algorithm not only leverages the backward (class) information flow from the output layer(s) to the input, but also the the forward (image) information extracted at various layers of abstraction. See Figure~\ref{fig:main}. 

\vspace{-7pt}
More specifically, our method has the following distinguishing characteristics:

\begin{enumerate}[nosep,leftmargin=*]
\item We propose a mathematically principled approach to achieve ``backward information flow'' within a deep convolutional network, leveraging the ideas proposed in the deconvolutional networks approach of~\cite{deconvnet}. However, this approach is computationally very expensive since it requires solving a sparse recovery problem for each layer of the network, and this limits the depth of a network on which this method is applicable. On the other hand, our approach only needs simple application of matrix adjoints and (element-wise) nonlinearities for each convolutional layer and can be easily implementable on very deep networks.
\item We propose a systematic way of using the forward information to guide the backward-traversal. In particular, we use the forward information to extract a support within a given layer of representation that best corresponds to a specific feature map. We achieve this using a novel \emph{masking} scheme which transparently combines both forward and backward information flows through the network.
\item As opposed to gradient-based schemes (such as~\cite{gradcam,guided-backprop}) that aggregate the information from all feature maps, our algorithm produces binary support estimates layer by layer. In that sense, our method avoids any numerical stability and robustness issues that may arise via the well-known problem of exploding/vanishing gradients that can potentially affect the interpretability. In particular, in contrast with \cite{deeplift}, we remove the need for any separate reference image, and our method only involves making two passes through the network for a given image.
\end{enumerate}

We present preliminary numerical evidence supporting our method, and demonstrate advantages over gradient-based methods such as guided backpropagation~\cite{guided-backprop}. 

\section{Proposed Approach: Forward-Backward Interpretability}



We now describe our scheme for visualizing a
convolutional neural network. We term our method \emph{Forward-Backward Interpretability} (or FBI for short). Given a test image, the goal is to identify important regions that explain the prediction of the learned network. To do
this, we propagate the class-probed information back to the image pixel space
through the complete network, using the guidance of \emph{learned} model weights as well as the
\emph{forward} activations of each neuron in the network.

Our approach shares several similarities with the deconvolutional
networks (DeconvNet) approach introduced in \cite{deconvnet}. However,
instead of \emph{reconstructing} lower layer feature maps from higher
layer activations as in DeconvNet, we merely try to identify important
regions (supports) preserved in the forward activations in each layer from the backward (class-specific) information flow.


Suppose that we have already trained the network to an optimal state. In the forward pass, an image is presented to the network, and the activations in the entire network are computed. To explain the classification, we consider the class indicator
vector $\hat{y}$ where $\hat{y}_c = 1$ for the predicted class $c$ and zero otherwise,
and back-propagate this information to the input space. We use
$\hat{y}$ as the input of the
backward pass to approximately ``invert'' each layer, while iteratively filtering these inverses using the forward activations. The process is repeated until the input layer is reached.

\textbf{Dense layers.} For each fully-connected
layer (indexed by $l$), denote its activation as: 
$$z^{(l)} = W^{(l)}a^{(l-1)} +
b^{(l)},~a^{(l)} = ReLU(z^{(l)})$$ 
and the softmax activation 
$y = \sigma(a^{(L)})$ is achieved at the final ($L^{\textrm{th}}$) layer. Our goal is to traverse each of these layers backwards. In order to achieve this, we define the ``adjoint'' of each operation\footnote{The term ``adjoint'' is only loosely defined here due to the nonlinearities involved.} as follows. The adjoint of the softmax layer, $\hat{z}^{(L)}$, is defined point-wise such that 
$$\hat{z}_i =\begin{cases} 
\max(z^{(L)}) & i = c \\
\min(z^{(L)}) & i \neq c. 
\end{cases}$$ 
The adjoint of the ReLU activation function is the ReLU itself. Overall, the ``adjoint'' of each fully connected layer is $\hat{a}^{(l-1)} = (W^{(l)})^T(\hat{z}^{(l)} - b^{(l)})$.

\textbf{Foward-Backward masking.} The backward information flow is now filtered using the forward activations. Specifically, 
we only keep entries
of $\hat{a}^{(l)}$ such that their entry-wise product with respective
entries in $a^{(l)}$ are above some threshold parameter $\tau$. The other entries are
set to zero otherwise. This enables us to identify a candidate
\emph{support} corresponding to an interpretable feature in the input
of a given layer.

\textbf{Contributing feature maps}. Among many backward feature maps
at the top convolutional layer, we keep only $k$ of them in the
backward information flow. The contribution of each map is determined
as the total activation of the entire map. Hence, the features
irrelevant to the probed class are removed.

\textbf{Unpooling}. To perform an adjoint of the max pooling layer, we reshape the
obtained pooled map $\hat{a}^{(l)}_p$ from the backward pass and
replicate (copy) its values across the domain of the max
operator. Then, we evaluate entry-wise:
\[
  \hat{a}^{(l)}_{up} = \min(a^{(l)}_p, \hat{a}^{(l)}_p)
\]
This is similar to an analogous unpooling operation in DeconvNet; however, that approach only copies the value of $\hat{a}^{(l)}_p$ in a single location via switches that are stored in memory. In contrast, our new scheme allows the backward feature maps after unpooling to be not overly sparse, and retains enough spatial information about the interpretation. We note that the replication step is suitable for any downsampling filter of size $2 \times 2$ and stride 2 (wherein the receptive fields are non-overlapping). For other filter sizes and stride lengths, the replicated values are averaged over the overlapping locations.

\textbf{Deconvolution}. The deconvolution step is similar to DeconvNet, where we compute the adjoint by convolving the backward activation with the flipped filter weights of a corresponding filter.

As we traverse backwards through the network using the above operations, we successively retain a subset of pixel indices (or support) at the input of each layer that plausibly corresponds to the ``interpretable'' portion of the given image. In the end, we display the locations of these indices together with the values produced by the adjoint. The selectivity achieved by successive masking means that we always obtain a fairly sparse support in our final estimate; the sparsity can be controlled via appropriate choice of the threshold parameter $\tau$. 





\section{Results}

We visualize the interpretations provided by the proposed algorithm in Table~\ref{Tab:FBI}. We use VGG-16 model pretrained with ImageNet dataset~\cite{chollet2015keras}. The visualizations are generated for the top 1 predictions of the image. The choice of top $k$ filters for computing the inverse is an important factor contributing to the interpretation obtained. If the value of $k$ is very less (around 10\% of the filters), then the interpretations obtained looses a lot of important features and when $k$ is too high (close to 100\%) then we see that the interpretations are too noisy. We have found that the visualizations obtained while using 100\% of the filters, and no pointwise thresholding based on the forward function value, produces an output that is close to the DeconvNet algorithm. Thus, using the forward function and masking the inverse computed as explained above, we achieve better output than the DeconvNet as well as the guided backpropagation algorithms.

For the experiments shown in the Table~\ref{Tab:FBI}, we use the top 50\% filters to propagate the inverse of each layer. We also use a thresholding value of 10.0 for the pointwise mask between the forward and backward values. We compare our method with the guided backpropagation algorithm. We notice that the resulting visualizations have lesser noise compared to that of the guided backpropagation algorithm.

\begin{table}[!t]
	\caption{Illustrative examples of the forward-backward inversion visualizations.}\vspace{0.1in}
	\label{Tab:FBI}
	\centering
	\small
	\newcommand\T{\rule{0pt}{2.7ex}}
	\newcommand\B{\rule[-1.3ex]{0pt}{0pt}}
	\newcommand{\tabincell}[2]{\begin{tabular}{@{}#1@{}}#2\end{tabular}}
	\tymin=.1in
	\tymax=2.5in 
	\begin{tabular}{lcccccc}
		\hline
		Input&\tabincell{c}{\\\includegraphics[width=0.15\linewidth,height=0.15\linewidth]{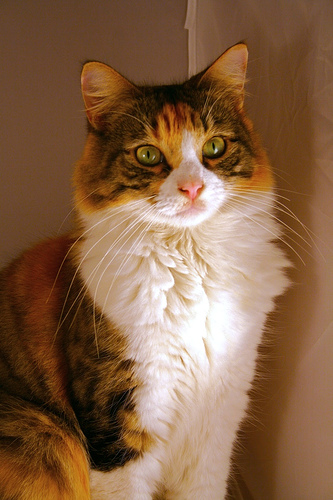}}&
		\tabincell{c}{\\\includegraphics[width=0.17\linewidth]{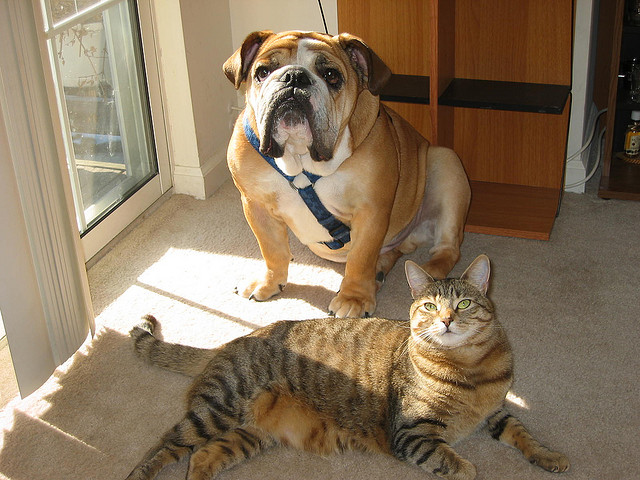}}&
		\tabincell{c}{\\\includegraphics[width=0.17\linewidth]{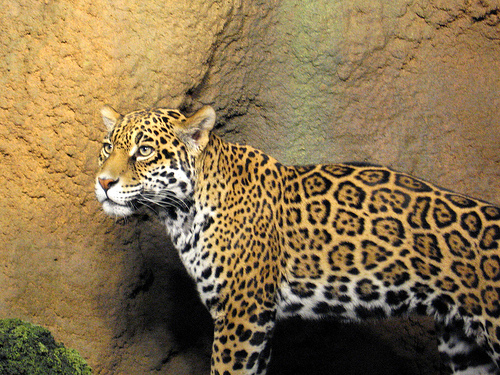}}&
		\tabincell{c}{\\\includegraphics[width=0.17\linewidth]{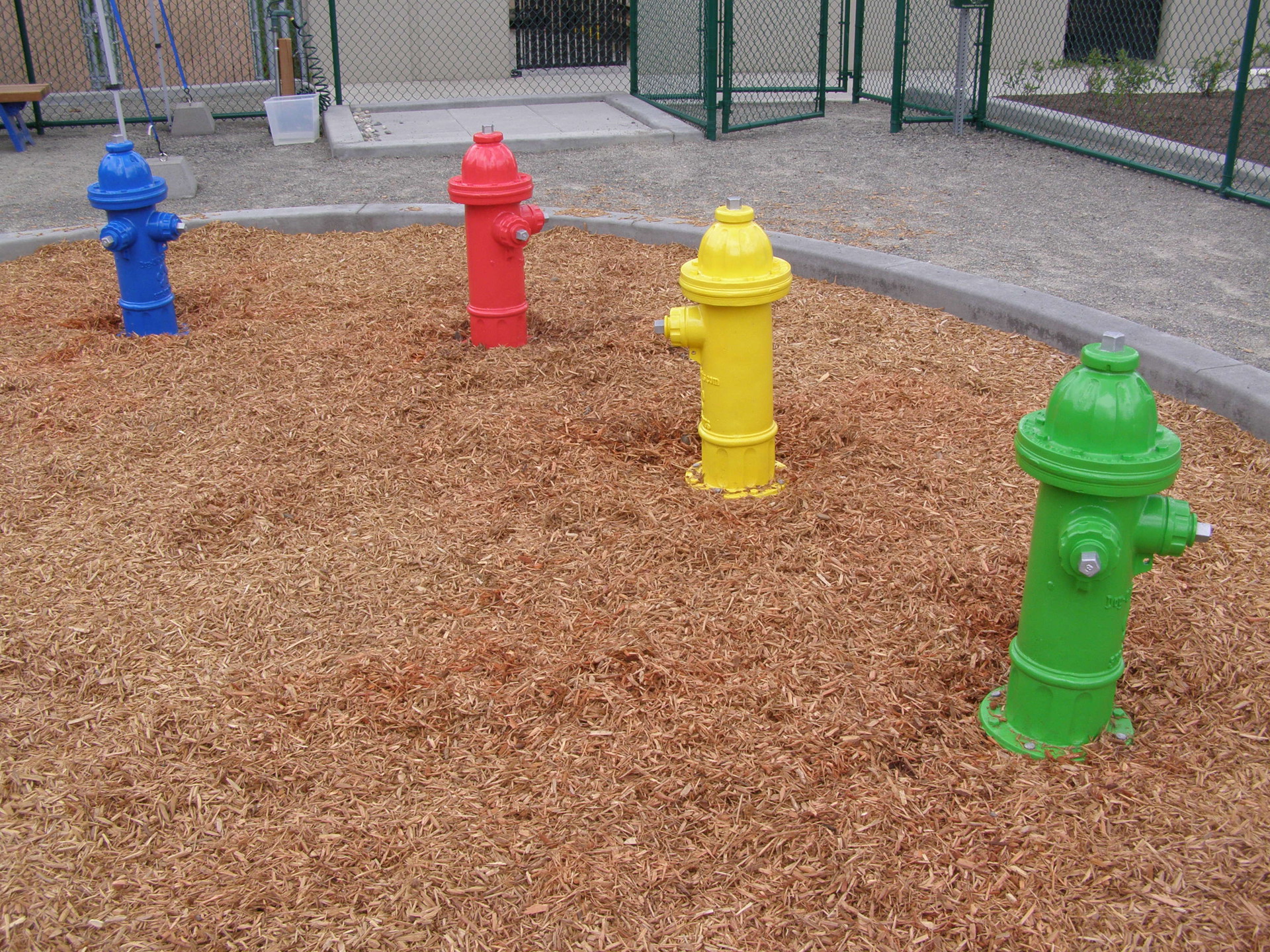}} 
		\\
		\hline                
		\tabincell{l}{\\FBI\\Interpretations\\ \{our method\}}& 
		\tabincell{c}{\\\includegraphics[width=0.17\linewidth]{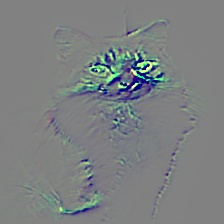}\\tabby cat}&
		\tabincell{c}{\\\includegraphics[width=0.17\linewidth]{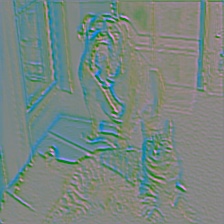}\\boxer}&
		\tabincell{c}{\\\includegraphics[width=0.17\linewidth]{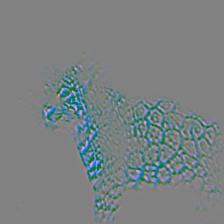}\\jaguar}&
		\tabincell{c}{\\\includegraphics[width=0.17\linewidth]{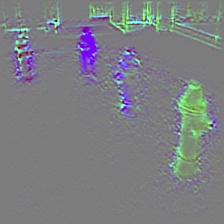}\\fire hydrant}\\
		\hline
		\tabincell{l}{\\Guided Backprop\\Visualizations}& 
		\tabincell{c}{\\\includegraphics[width=0.17\linewidth]{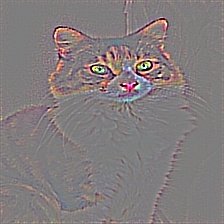}\\tabby cat}&
		\tabincell{c}{\\\includegraphics[width=0.17\linewidth]{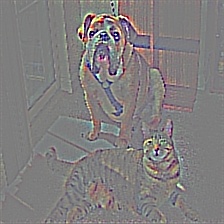}\\boxer}&
		\tabincell{c}{\\\includegraphics[width=0.17\linewidth]{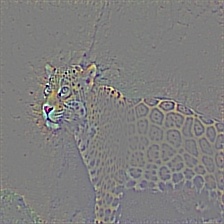}\\jaguar}&
		\tabincell{c}{\\\includegraphics[width=0.17\linewidth]{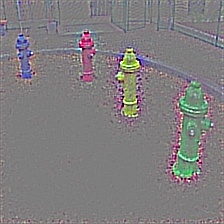}\\fire hydrant}\\
		\hline                
		
	\end{tabular}
\end{table}

\section{Conclusions}

In this work, we introduce a novel Forward-Backward approach for visualizing the interpretations that correspond to a particular class. However, we see that the choice of $k$ filters for computing the interpretations has become a hyper-parameter to ensure that the interpretations are good. It is also seen that the filters which contribute to one particular class might have similar activations to the activations pertaining to some other class. Hence, decoupling the activations of these filters to choose which filters to use for computing the inverse, so that we maintain class discriminativeness, is something of much interest to the authors.

%

{\small
\bibliographystyle{unsrtnat}
\bibliography{NIPS_SYMP/allrefs}
}

\end{document}